\documentclass[11pt,a4paper]{article}
\usepackage[hyperref]{acl2017}
\usepackage{times}
\usepackage{latexsym}
\usepackage{amsmath}
\usepackage{diagbox}
\usepackage{textcomp}
\usepackage[super]{nth}

\aclfinalcopy 

\title{CompiLIG at SemEval-2017 Task 1: Cross-Language Plagiarism Detection Methods for Semantic Textual Similarity}

\author{
  J\'{e}r\'{e}my Ferrero \\
  Compilatio \\
  276 rue du Mont Blanc \\
  74540 Saint-F\'{e}lix, France \\
  LIG-GETALP \\
  Univ. Grenoble Alpes, France \\
  {\tt jeremy.ferrero@imag.fr} \\
  \And
  Fr\'{e}d\'{e}ric Agn\`{e}s \\
  Compilatio \\
  276 rue du Mont Blanc \\
  74540 Saint-F\'{e}lix, France \\
  {\tt frederic@compilatio.net} \\
  \AND
  Laurent Besacier \\
  LIG-GETALP \\
  Univ. Grenoble Alpes, France \\
  {\tt laurent.besacier@imag.fr} \\
  \And
  Didier Schwab \\
  LIG-GETALP \\
  Univ. Grenoble Alpes, France \\
  {\tt didier.schwab@imag.fr} \\
}

\date{}

\begin{document}
\maketitle
\begin{abstract}

We present our submitted systems for Semantic Textual Similarity (STS) Track~4 at SemEval-2017.
Given a pair of Spanish-English sentences, each system must estimate their semantic similarity by a score between~0 and~5.
In our submission, we use syntax-based, dictionary-based, context-based, and MT-based methods.
We also combine these methods in unsupervised and supervised way.
Our best run ranked \nth{1} on track~4a with a correlation of~83.02\% with human annotations.

\end{abstract}

\section{Introduction}
\label{intro}

\mbox{CompiLIG} is a collaboration between \mbox{Compilatio}\footnote{\url{www.compilatio.net}} - a company particularly interested in cross-language plagiarism detection - and \mbox{LIG} research group on natural language processing (\mbox{GETALP}).
Cross-language semantic textual similarity detection is an important step for cross-language plagiarism detection, and evaluation campaigns in this new domain are rare.
For the first time, SemEval STS task \cite{semeval2016} was extended with a Spanish-English cross-lingual sub-task in 2016.
This year, sub-task was renewed under track~4 (divided in two sub-corpora: track~4a and track~4b).

Given a sentence in Spanish and a sentence in English, the objective is to compute their semantic textual similarity according to a score from~0 to~5, where~0 means \emph{no similarity} and~5 means \emph{full semantic similarity}.
The evaluation metric is a \mbox{Pearson} correlation coefficient between the submitted scores and the gold standard scores from human annotators. 
Last year, among 26 submissions from~10 teams, the method that achieved the best performance \cite{uwb} was a supervised system (SVM regression with RBF kernel) based on word alignment algorithm presented in \newcite{sultan}.


\textbf{Our submission} in 2017 is based on cross-language plagiarism detection methods combined with the best performing STS detection method published in 2016.
\mbox{CompiLIG} team participated to SemEval STS for the first time in 2017.
The methods proposed are syntax-based, dictionary-based, context-based, and MT-based.
They show additive value when combined.
The submitted runs consist in (1) our best single unsupervised approach (2) an unsupervised combination of best approaches (3) a fine-tuned combination of best approaches.
The best of our three runs ranked \nth{1} with a correlation of~83.02\% with human annotations on track~4a among all submitted systems (51 submissions from~20 teams for this track).
Correlation results of all participants (including ours) on track~4b were much lower and we try to explain why (and question the validity of track~4b) in the last part of this paper.

\section{Cross-Language Textual Similarity Detection Methods}
\label{methods}

\subsection{Cross-Language Character \mbox{N-Gram} (CL-C$n$G)}
\label{cl-cng}
\emph{\mbox{CL-C$n$G}} aims to measure the syntactical similarity between two texts.
It is based on \newcite{mcnamee2004} work used in information retrieval.
It compares two texts under their n-grams vectors representation.
The main advantage of this kind of method is that it does not require any translation between source and target text.

After some tests on previous year's dataset to find the best $n$, we decide to use the \newcite{potthast2011}'s \mbox{\emph{CL-C$3$G}} implementation.
Let~$S_{x}$ and~$S_{y}$ two sentences in two different languages.
First, the alphabet of these sentences is normalized to the ensemble $\sum\,=\,\{a-z, 0-9,\quad\}$, so only spaces and alphanumeric characters are kept.
Any other diacritic or symbol is deleted and the whole text is lower-cased.
The texts are then segmented into~3-grams (sequences of~3 contiguous characters) and transformed into \emph{\mbox{tf.idf}} vectors of character~3-grams.
We directly build our \emph{\mbox{idf}} model on the evaluation data.
We use a double normalization \emph{K} (with \emph{K}\,=\,0.5) as \emph{\mbox{tf}} \cite{tfidf_manning2008} and a inverse document frequency smooth as \emph{\mbox{idf}}.
Finally, a cosine similarity is computed between the vectors of source and target sentences.

\subsection{Cross-Language Conceptual Thesaurus-based Similarity (CL-CTS)}
\label{cl-cts}
\emph{\mbox{CL-CTS}} \cite{gupta2012, pataki2012} aims to measure the semantic similarity between two vectors of concepts. 
The model consists in representing texts as bag of words (or concepts) to compare them. 
The method also does not require explicit translation since the matching is performed using internal connections in the used ``ontology".

Let~$S$ a sentence of length~$n$, the~$n$ words of the sentence are represented by~$w_{i}$ as:
\begin{equation} 
S = \{w_{1}, w_{2}, w_{3}, ..., w_{n}\} 
\end{equation}

$S_{x}$ and~$S_{y}$ are two sentences in two different languages. 
A bag of words~$S'$ from each sentence~$S$ is built, by filtering stop words and by using a function that returns for a given word all its possible translations. 
These translations are jointly given by a linked lexical resource, \emph{\mbox{DBNary}} \cite{dbnary}, and by cross-lingual word embeddings.
More precisely, we use the top~10 closest words in the embeddings model and all the available translations from \emph{\mbox{DBNary}} to build the bag of words of a word.
We use the \emph{\mbox{MultiVec}} \cite{multivec} toolkit for computing and managing word embeddings.
The corpora used to build the embeddings are Europarl and Wikipedia sub-corpus, part of the dataset of \newcite{dataset-lrec}\footnote{\url{https://github.com/FerreroJeremy/Cross-Language-Dataset}}.
For training our embeddings, we use CBOW model with a vector size of~100, a window size of~5, a negative sampling parameter of~5, and an alpha of~0.02.

So, the sets of words~$S'_{x}$ and~$S'_{y}$ are the conceptual representations in the same language of~$S_{x}$ and~$S_{y}$ respectively.
To calculate the similarity between~$S_{x}$ and~$S_{y}$, we use a syntactically and frequentially weighted augmentation of the Jaccard distance, defined as:
\begin{equation}
J(S_{x}, S_{y}) = \dfrac{\Omega(S'_{x}) + \Omega(S'_{y})}{\Omega(S_{x}) + \Omega(S_{y})}
\end{equation}

where~$S_{x}$ and~$S_{y}$ are the input sentences (also represented as sets of words), and~$\Omega$ is the sum of the weights of the words of a set, defined as:
\begin{equation}
\Omega(S) = \sum_{i=1\,,\,w_{i} \in S}^{n}\varphi(w_{i})
\end{equation}

where~$w_{i}$ is the~$i^{th}$ word of the bag~$S$, and~$\varphi$ is the weight of word in the Jaccard distance:
\begin{equation}
\varphi(w) = pos\_weight(w)^{1-\alpha}\ .\ idf(w)^{\alpha}
\end{equation}

where~$pos\_weight$ is the function which gives the weight for each universal part-of-speech tag of a word, $idf$ is the function which gives the inverse document frequency of a word, and~$.$~is the scalar product.
Equation (4) is a way to syntactically ($pos\_weight$) and frequentially ($idf$) weight the contribution of a word to the Jaccard distance (both contributions being controlled with the $\alpha$ parameter).
We assume that for one word, we have its part-of-speech within its original sentence, and its inverse document frequency.
We use \emph{\mbox{TreeTagger}} \cite{schmid1994} for POS tagging, and we normalize the tags with \emph{Universal Tagset} of \newcite{universal-tagset}.
Then, we assign a weight for each of the~12 universal POS tags.
The 12 POS weights and the value~$\alpha$ are optimized with \emph{\mbox{Condor}} \cite{condor} in the same way as in \newcite{clwes2017}.
\emph{\mbox{Condor}} applies a Newton\textquotesingle s method with a trust region algorithm to determinate the weights that optimize a desired output score.
No re-tuning of these hyper-parameters for SemEval task was performed.

\subsection{Cross-Language Word Embedding-based Similarity}
\label{cl-wes}

\emph{\mbox{CL-WES}} \cite{clwes2017} consists in a cosine similarity on distributed representations of sentences, which are obtained by the weighted sum of each word vector in a sentence.
As in previous section, each word vector is syntactically and frequentially weighted.

If~$S_{x}$ and~$S_{y}$ are two sentences in two different languages, then \emph{\mbox{CL-WES}} builds their (bilingual) common representation vectors~$V_{x}$ and~$V_{y}$ and applies a cosine similarity between them.
A distributed representation~$V$ of a sentence~$S$ is calculated as follows:
\begin{equation}
V = \sum_{i=1\,,\,w_{i} \in S}^{n}(vector(w_{i})\,.\,\varphi(w_{i}))
\end{equation}

where~$w_{i}$ is the~$i^{th}$ word of the sentence~$S$, $vector$ is the function which gives the word embedding vector of a word, $\varphi$ is the same that in formula~(4), and~$.$~is the scalar product.
We make this method publicly available through \emph{\mbox{MultiVec}}\footnote{\url{https://github.com/eske/multivec}} \cite{multivec} toolkit.

\subsection{Translation + Monolingual Word Alignment (T+WA)}
\label{t+wa}

The last method used is a two-step process.
First, we translate the Spanish sentence into English with \emph{\mbox{Google Translate}} (\textit{i.e.} we are bringing the two sentences in the same language).
Then, we align both utterances.
We reuse the monolingual aligner\footnote{\url{https://github.com/ma-sultan/monolingual-word-aligner}} of \newcite{sultan} with the improvement of \newcite{uwb}, who won the cross-lingual sub-task in 2016 \cite{semeval2016}.
Because this improvement has not been released by the initial authors, we propose to share our re-implementation on \emph{GitHub}\footnote{\url{https://github.com/FerreroJeremy/monolingual-word-aligner}}.

If~$S_{x}$ and~$S_{y}$ are two sentences in the same language, then we try to measure their similarity with the following formula:
\begin{equation}
J(S_{x}, S_{y}) = \dfrac{\omega(A_{x}) + \omega(A_{y})}{\omega(S_{x}) + \omega(S_{y})}
\end{equation}

where~$S_{x}$ and~$S_{y}$ are the input sentences (represented as sets of words), $A_{x}$ and~$A_{y}$ are the sets of aligned words for~$S_{x}$ and~$S_{y}$ respectively, and~$\omega$ is a frequency weight of a set of words, defined as:
\begin{equation}
\omega(A) = \sum_{i=1\,,\,w_{i} \in A}^{n} idf(w_{i})
\end{equation}

where~$idf$ is the function which gives the inverse document frequency of a word.

\subsection{System Combination}
\label{fusion}

These methods are syntax-, dictionary-, context- and MT- based, and are thus potentially complementary.
That is why we also combine them in unsupervised and supervised fashion.
Our unsupervised fusion is an average of the outputs of each method.
For supervised fusion, we recast fusion as a regression problem and we experiment all available methods in \emph{\mbox{Weka}}~3.8.0 \cite{weka}.

\section{Results on SemEval-2016 Dataset}
\label{results2016}

Table~\ref{results2016table} reports the results of the proposed systems on SemEval-2016 STS cross-lingual evaluation dataset.
The dataset, the annotation and the evaluation systems were presented in the SemEval-2016 STS task description paper \cite{semeval2016}, so we do not re-detail them here.
The lines in bold represent the methods that obtain the best mean score in each category of system (best method alone, unsupervised and supervised fusion).
The scores for the supervised systems are obtained with 10-folds cross-validation.

\begin{table}[th!]
\begin{center}
\begin{small}
\begin{tabular}{|l|l|l|l|}
      \hline
      \bf Methods & \bf News & \bf Multi & \bf Mean \\
      \hline \hline
      \multicolumn{4}{|c|}{Unsupervised systems} \\
      \hline \hline
      CL-C$3$G (1) & 0.7522 & 0.6550 & 0.7042 \\
      \textbf{CL-CTS (2)} & \textbf{0.9072} & \textbf{0.8283} & \textbf{0.8682} \\
      CL-WES (3) & 0.7028 & 0.6312 & 0.6674 \\
      T+WA (4) & 0.9060 & 0.8144 & 0.8607 \\
      \hline \hline
      Average (1-2-3-4) & 0.8589 & 0.7824 & 0.8211 \\
      \textbf{Average (1-2-4)} & \textbf{0.9051} & \textbf{0.8347} & \textbf{0.8703} \\
      Average (2-3-4) & 0.8923 & 0.8239 & 0.8585 \\
      Average (2-4) & 0.9082 & 0.8299 & 0.8695 \\
      \hline \hline
      \multicolumn{4}{|c|}{Supervised systems (fine-tuned fusion)} \\
      \hline \hline
      GaussianProcesses & 0.8712 & 0.7884 & 0.8303 \\
      LinearRegression & 0.9099 & 0.8414 & 0.8761 \\
      MultilayerPerceptron & 0.8966 & 0.7999 & 0.8488 \\
      SimpleLinearRegression & 0.9048	 & 0.8144 & 0.8601 \\
      SMOreg & 0.9071 & 0.8375 & 0.8727 \\
      Ibk & 0.8396 & 0.7330	 & 0.7869 \\
      Kstar & 0.8545 & 0.8173 & 0.8361 \\
      LWL & 0.8572 & 0.7589 & 0.8086 \\
      DecisionTable & 0.9139 & 0.8047 & 0.8599 \\
      M5Rules & 0.9146	 & 0.8406 & 0.8780 \\
      DecisionStump & 0.8329 & 0.7380 & 0.7860 \\
      \textbf{M5P} & \textbf{0.9154} & \textbf{0.8442} & \textbf{0.8802} \\
      RandomForest & 0.9109 & 0.8418 & 0.8768 \\
      RandomTree & 0.8364 & 0.7262 & 0.7819 \\
      REPTree & 0.8972 & 0.7992 & 0.8488 \\
      \hline
\end{tabular}
\end{small}
\end{center}
\caption{\label{results2016table} Results of the methods on SemEval-2016 STS cross-lingual evaluation dataset.}
\end{table}

\section{Runs Submitted to SemEval-2017}
\label{submissions}
First, it is important to mention that our outputs are linearly re-scaled to a real-valued space~[0\,;\,5].

\textbf{Run 1: Best Method Alone.}
\label{solo-method}
Our first run is only based on the best method alone during our tests (see Table~\ref{results2016table}), \textit{i.e.} Cross-Language Conceptual Thesaurus-based Similarity (\emph{\mbox{CL-CTS}}) model, as described in section~\ref{cl-cts}.

\textbf{Run 2: Fusion by Average.}
\label{average}
Our second run is a fusion by average on three methods: \emph{\mbox{CL-C$3$G}}, \emph{\mbox{CL-CTS}} and \emph{\mbox{T+WA}}, all described in section~\ref{methods}.

\textbf{Run 3: M5$'$ Model Tree.}
\label{m5p}
Unlike the two precedent runs, the third run is a supervised system.
We have selected the system that obtained the best score during our tests on SemEval-2016 evaluation dataset (see Table~\ref{results2016table}), which is the M5$'$~model tree \cite{m5p} (called M5P in \emph{\mbox{Weka}}~3.8.0 \cite{weka}).
Model trees have a conventional decision tree structure but use linear regression functions at the leaves instead of discrete class labels.
The first implementation of model trees, M5, was proposed by \newcite{m5} and the approach was refined and improved in a system called M5$'$~by \newcite{m5p}.
To learn the model, we use all the methods described in section~\ref{methods} as features.

\section{Results of the 2017 evaluation and Discussion}
\label{results2017}

Dataset, annotation and evaluation systems are presented in SemEval-2017 STS task description paper \cite{semeval2017}.
We can see in Table~\ref{results2017table} that our systems work well on SNLI\footnote{\url{http://nlp.stanford.edu/projects/snli/}} \cite{snli} (track~4a), on which we ranked \nth{1} with more than~83\% of correlation with human annotations.
Conversely, correlations on the WMT corpus (track~4b) are strangely low.
This difference is notable on the scores of all participating teams \cite{semeval2017}\footnote{The best score for this track is 34\%, while for the other tracks it is around 85\%.}.
This might be explained by the fact that WMT was annotated by only one annotator, while the SNLI corpus was annotated by many.

\begin{table}[th!]
\begin{center}
\begin{small}
\begin{tabular}{|l|l|l|l|}
      \hline
      \bf Methods & \bf SNLI (4a) & \bf WMT (4b) & \bf Mean \\
      \hline \hline
      CL-CTS & 0.7684 & 0.1464 & 0.4574 \\
      Average & 0.7910 & 0.1494 & 0.4702 \\
      M5P & 0.8302 & 0.1550 & 0.4926 \\
      \hline
\end{tabular}
\end{small}
\end{center}
\caption{\label{results2017table} Official results of our submitted systems on SemEval-2017 STS track~4 evaluation dataset.}
\end{table}

\begin{table}[th!]
\begin{center}
\begin{small}
\begin{tabular}{|l|l|l|l|}
      \hline
      \bf Methods & \bf SNLI (4a) & \bf WMT (4b) & \bf Mean \\
      \hline
      \hline
      \multicolumn{4}{|c|}{Our Annotations} \\
      \hline
      \hline
      CL-CTS & 0.7981 & 0.5248 & 0.6614 \\
      Average & 0.8105 & 0.4031 & 0.6068 \\
      M5P & 0.8622 & 0.5374 & 0.6998 \\
      \hline
      \hline
      \multicolumn{4}{|c|}{SemEval Gold Standard} \\
      \hline
      \hline
      CL-CTS & 0.8123 & 0.1739 & 0.4931 \\
      Average & 0.8277 & 0.2209 & 0.5243 \\
      M5P & 0.8536 & 0.1706 & 0.5121 \\
      \hline
\end{tabular}
\end{small}
\end{center}
\caption{\label{methods_on_our_annotations} Results of our submitted systems scored on our 120 annotated pairs and on the same 120 SemEval annotated pairs.}
\end{table}

To investigate deeper on this issue, we manually annotated 60 random pairs of each sub-corpus (120 annotated pairs among 500).
These annotations provide a second annotator reference.
We can see in Table~\ref{methods_on_our_annotations} that, on SNLI corpus (4a), our methods behave the same way for both annotations (a difference of about 1.3\%).
However, the difference in correlation is huge between our annotations and SemEval gold standard on the WMT corpus (4b): 30\% on average.
The \mbox{Pearson} correlation between our annotated pairs and the related gold standard is 85.76\% for the SNLI corpus and 29.16\% for the WMT corpus.
These results question the validity of the WMT corpus (4b) for semantic textual similarity detection.

\section{Conclusion}
\label{conclusion}

We described our submission to SemEval-2017 Semantic Textual Similarity task on track~4 (Sp-En cross-lingual sub-task).
Our best results were achieved by a M5$'$~model tree combination of various textual similarity detection techniques.
This approach worked well on the SNLI corpus (4a - finishes \nth{1} with more than~83\% of correlation with human annotations), which corresponds to a real cross-language plagiarism detection scenario. We also questioned WMT corpus (4b) validity providing our own manual annotations and showing low correlations with those of SemEval.

\bibliography{acl2017}
\bibliographystyle{acl_natbib}

\appendix

\end{document}